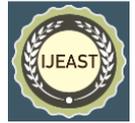

# CRIMINAL INVESTIGATION TRACKER WITH SUSPECT PREDICTION USING MACHINE LEARNING

Dilmini S. J
Faculty of Computing
Sri Lanka Institute of Information Technology
Malabe, Sri Lanka

Rajapaksha R.A.T.M
Faculty of Computing
Sri Lanka Institute of Information Technology
Malabe, Sri Lanka

Erandika Lakmali
University of Kelaniya
Dalugama, Kelaniya, Sri Lanka

Mandula S.P. S
Faculty of Computing
Sri Lanka Institute of Information Technology
Malabe, Sri Lanka

Delgasdeniya D.D. G
Faculty of Computing
Sri Lanka Institute of Information Technology
Malabe, Sri Lanka

Pradeepa Bandara
Faculty of Computing
Sri Lanka Institute of Information Technology
Malabe, Sri Lanka

*Abstract* - **An automated approach to identifying offenders in Sri Lanka would be better than the current system. Obtaining information from eyewitnesses is one of the less reliable approaches and procedures still in use today. Automated criminal identification has the ability to save lives, notwithstanding Sri Lankan culture's lack of awareness of the issue. Using cutting-edge technology like biometrics to finish this task would be the most accurate strategy. The most notable outcomes will be obtained by applying fingerprint and face recognition as biometric techniques. The main responsibilities will be image optimization and criminality. CCTV footage may be used to identify a person's fingerprint, identify a person's face, and identify crimes involving weapons. Additionally, we unveil a notification system and condense the police report to Additionally, to make it simpler for police officers to understand the essential points of the crime, we develop a notification system and condense the police report. Additionally, if an incident involving a weapon is detected, an automated notice of the crime with all the relevant facts is sent to the closest police station. The summarization of the police report is what makes this the most original. In order to improve the efficacy of the overall image, the system will quickly and precisely identify the full crime scene, identify, and recognize the suspects using their faces and fingerprints, and detect firearms. This study provides a novel approach for crime prediction based on real-world data, and criminality incorporation. A crime or occurrence should be reported to the appropriate agencies, and the suggested web application should be improved further to offer a workable channel of communication.**

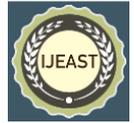

## I. INTRODUCTION

This research presents a desktop program that is more efficient than current approaches for quickly and accurately identifying offenders based on suspects' faces, fingerprints, and the detection of weapons. There is presently no comprehensive system in place in Sri Lanka that makes use of cutting-edge technology to spot illegal activity and alert the appropriate authorities right away. A system will acquire and preserve information on crimes while taking the circumstances into account, ensuring that parties and the rest of the community are aware of specific occurrences in real time. It will then look at and add information that could be useful for future research. A team must come up with and implement a sophisticated approach for compiling data and analyzing it to get the appropriate conclusion in line with that vast field of research. The Sri Lankan Penal Code was initially adopted as a reaction to the ambiguity and sufficiency of the nation's pre-existing criminal legislation. Despite some updates, this Penal Code is still in effect today. It is recognized that insanity, drunkenness, need, duress, and private defense are all strong defenses against criminal liability. The ability to reason, or "men's rea," is deemed absent in children under the age of eight [1].

A crime is defined as "an act done in violation of a law banning it or omitted in violation of a law ordering it" in its most basic form. Anyone who commits a crime may be subject to repercussions from the police and the government since it is a disgusting and illegal activity that breaks the law [2]. A criminal is a person who has either committed or is currently committing a crime. Crime is a societal blight that has a wide range of negative effects on our society. Criminality is growing increasingly pervasive in our society. 10% of criminals are responsible for 50% of all crimes. Crime is essentially an "unpredictable" occurrence. It is not limited by space or time. Everything depends on what individuals do. Illegal actions can take many different forms, ranging from careless driving to terrorist attacks. Criminals produce a considerable quantity of data via their various operations, which may be accessible in a range of ways. As a result, it might be challenging to assess crime data. Data mining is a method for obtaining useful information from enormous amounts of data [3]. Technology is underutilized when it comes to deterring crime and identifying offenders, while criminals today employ more advanced tools to perpetrate crimes.

## II. RELATED WORK

The same subjects have been emphasized within the literature review, along with the related research roles and components. The major goal of this research is to create a completely automated criminal investigation and tracking system in Sri Lanka that includes a suspect prediction system using CCTV images of public places. This central system is composed of four components. They are the following: the face detection system, the weapon detection, and alarm trigging systems, the fingerprint detection system, the smart notification system, and the content summarizer. There isn't a system in operation right now that has all four of these components.

Algorithms for detecting and identifying fingerprints in criminal investigations were developed by Khin Nandar Win and his colleagues. Fingerprints are the skin's identifying features. We may use it to identify someone because of its unusual roughness and form. Additionally, fingerprints can be created by applying fluids, such as ink or plasma, to the skin or by pressing an assumed print into a delicate layer [4]. The fingerprint image's quality has to be raised in order to boost biometric identification's reliability when used with AFIS. Ding et al. developed the orientation-selective 2D Adaptive Chebyshev Band-pass Filter (ACBF), also known as the fingerprint enhancement filter [5].

Wati and his colleagues demonstrated a system design for Face Detection and Recognition in Smart Home Security. They also discussed how the concept was implemented using MyRIO 1900 and LabVIEW, respectively. A camera that is USB-connected to MyRIO is used to take pictures of people. When tested with a live picture, the facial recognition technology attained an accuracy of 80% [6]. The most common form of home security system requires constant use of the key to open and close the door and is mechanical. It is believed that using traditional security methods is ineffective and wasteful. Using the face increases security because it cannot be reproduced or changed hands [7].

The closed-circuit television (CCTV) operator develops video blindness after 20 to 40 minutes of active monitoring, according to a study by Dadashi and his colleagues (Research, 2003; Cohen et al., 2009; Dadashi, 2008). Over the past two decades, academics and industry professionals have focused their research on developing surveillance systems that can detect suspicious conduct (Zhou & Tan, 2010; Liwei et al., 2010; Kishore et al., 2012; Mandrupkar et al., 2013). Automation is required for complex scenarios to reduce the effort of the human operator and improve performance [8]. As a result, it is still required to intervene, modernize, and switch from traditional to intelligent and smart monitoring systems (Shah et al., 2007; Tian et al., 2008). There is ZERO human participation in IVSS. The sophisticated monitoring system immediately sends out a warning in the event of any suspicious behavior or illegal activity. The operator is able to focus completely on the video feed and take useful action as a consequence.

S. A. Kovalchik and his team looked at new measures that provide metrics for an information-sharing system in a community of sex offenders in Southern California. On each of the four focus occurrences, offenders were matched (citations, field interviews, crime cases, and arrests). The quantity and timing of police events were related to the sharing of information on sex offender registry status [9].

Text summarizing, which Laith Abualigah and his four collaborators devised, creates summaries from raw sources by retaining noteworthy information. Systems for text summarizing follow extractive and abstractive principles. They create extractive-compressive summaries for oracles. KALIMAT corpus of 18 M words is utilized to assess the proposed method [10].

### III. RESEARCH OBJECTIVES

Through public place monitoring, this research aims to identify crimes and provide the essential information and a visual representation of the reported occurrence to the appropriate security services. This project aims to find and identify offenders more quickly and precisely than the present methods by using their faces, fingerprints, weapons, and picture enhancement. Additionally, the police report summary is presented here. The sub-goals are as follows:

*A.* Fingerprint images are used to identify suspects. Those images might not be appropriate for use in the following procedures. A high-quality picture is necessary in order for the system to identify fingerprints with accuracy. It is required to increase the matching rate of the image enhancement method's processing speed and quality in comparison to other techniques on the market.

*B.* The core objective of developing an AI-based solution to assess obtained data and provide them as crucial proof comes to mind when it comes to the approach. As a result, the main emphasis should be on real-time video collection, processing, and analysis of the data obtained, identification of criminal conduct, notification of the relevant authorities, and highlighting of the original footage.

*C.* Creating and deploying an automated system to detect and identify weapon-related crimes and alarms are primarily focused on finding weapons like firearms and knives in public spaces that are being held by people to commit a damaging deed. Identify the most common weapons and crime trends here. Investigate and examine the data set, Create an appropriate algorithm for CCTV data preparation, Utilize an algorithm to determine the best method for feature extraction of firearms, Use live CCTV images, and train the model to identify and categorize firearms. Determine and train the model to quickly classify crimes involving weapons.

*D.* The content summarizer and smart notification system are the system's last stages. Because it is so helpful to the police department and other authorized parties in conducting a criminal investigation, this web-based application was created. Here, the notice will automatically send to the closest police station with the pertinent date, time, and position when the quiet alarm is activated at the spot where *it is* installed. The generation of the summary of the police report has also saved time in this case.

### III. METHODOLOGY

In this study, online apps are employed to identify suspects rapidly and reliably. A smart notification system, content summary, fingerprint detection, face identification, and object detection from the CCTV footage will conduct the main tasks alternately. Authorities can add offenders to the system thanks to our technology. Our technology is expected to be able to recognize and apprehend anyone who is intending to conduct or has already committed a crime. The proposed strategy provides Sri Lanka's state security sector great potential to curtail and ultimately eradicate criminal activity.

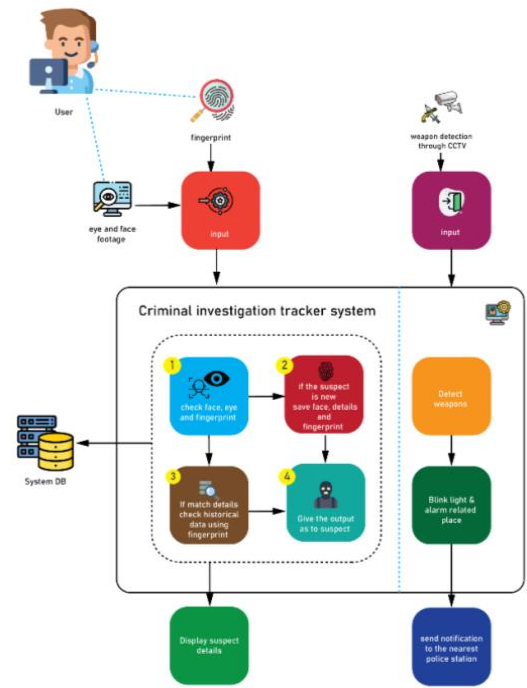

Fig. 1. Overall System Diagram

#### A. Fingerprint Detection -

First, we created a dataset and Oregon the images. and we build up a Siamese network with the images. the Siamese network is the core here which builds up with the convolutional neural network (CNN) architecture algorithm. it gives a similarity rate of 0,1 when we input the image of the fingerprint. Additionally, if a fingerprint is not already recorded in the database, one is created and added automatically.

#### B. Face detection and identification -

Here we used Face Net, a pre-trained model. we newly fine-tuned the pictures and after that when we give a new image to the system, it predicts by face-trained model and gives the ArcFace. later it saved this feature vector and gives a key. when we input a new suspect's face image and it checks the similarity between the face vector through cosine similarity. the Face Net is the main model and gives a 0,1 value via the cosine similarity matrix. it calculates the users' similarity when the rate is more than .8.

*C.* **Weapon detection and alarm trigging -**

Due to our interest in real-time detection, we are particularly concerned with classification speed. Additionally, we are particularly concerned with accuracy because a false alarm could result in very unfavorable reactions. These two factors make our project distinct from other Object Classification problems. When choosing the optimal course of action, we must find the ideal compromise between precision and quickness.

On the identification of firearms in picture and video data, some research has been done. We reviewed works on the detection of violent movie scenes, the analysis of infrared data for concealed weapons, and the identification of guns in body camera footage as part of our assessment of image classification research. Olmos, Tabik, and Herrera look at alarms that go off automatically when a gun is spotted in surveillance footage (Automatic Handgun Detection Alarm in Videos Using Deep Learning) [6]. Although this work uses deep learning/neural net techniques, the majority of research in the field has not. The project uses YOLO V7 to fine-tune the action sequence of a revolver being drawn.

*D.* **Smart notification system and content summarizer -**
The police can utilize a web application that is part of the smart notification system to be informed of crises and criminal activity. Here, we activate a quiet alarm in a crowded area (eg: Bank). If the system detects a weapon, the silent trigger activates and sends a notice directly to the nearest police station with the pertinent time, date, and position pinned on GPS.

The basic objective of text summarizing is to extract the most accurate and useful information from any data source. This situation is special in that Sri Lankan authorities have not yet discovered or shut down the police record and report converter. Summary creation is useful for many Natural Language Processing (NLP) tasks. In NLP, extractive and abstractive text summarization are the two basic categories. [11] I choose extractive summarization in this case since it draws out key terms from the provided text. The first step in the content summarization method used by our technology is to exclude frequent terms like am and are. After that, create an array from the sentence. Ex: I return home. I and go are often used terms. The highlight is home. The text is then numbered. The words are then given a numerical number and are added value in accordance with their significance. After that, create a NumPy array from the text. According to the word's and number's priority, a calculation is made. We then receive the finished product. To produce a word, that priority is required. It displays the text's strongest phrases.

With this, a sizable document is sent to the person who drafts the police report. This entails reading it to someone else while highlighting the key passages. They can read it effortlessly, which is a benefit. And all of this is done to shorten the time needed. When we provide a text or document, the result is significantly shorter than the one provided. This also serves as an introduction to Sri Lanka. This is frequently utilized while reading a police document and pulling out the key points. We used T5 architecture for Sinhala language summarization.

## IV. RESULTS AND DISCUSSION
This sector presents the results and discoveries obtained through training and testing methods.

*A.* **Fingerprint detection and identification system -**
The system compares and matches the picture that we enter with the database's image-matching rate. The algorithm recognizes him as a suspect if there is a matching rate of more than 80% from that. [12] The database then displays the suspect's information, including name, age, and ID card number. The database will save the supplied image as a new image if it does not match any of the photos already there.

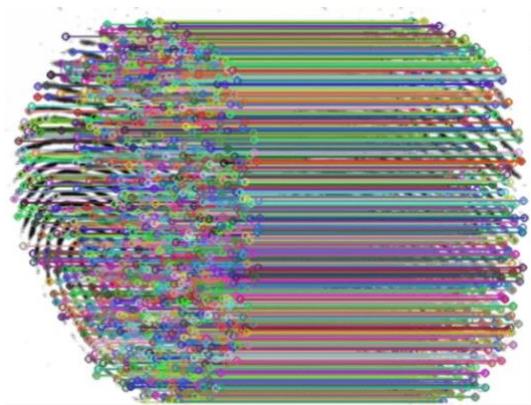

Fig. 2. Detection of fingerprint in python

*B.* **Face detection and identification system -**
Here, we discuss several facets of the face. This system has advanced sufficiently for it to distinguish a sideways-facing face. [13] We also become better to the point where we can recognize faces in hazy videos.

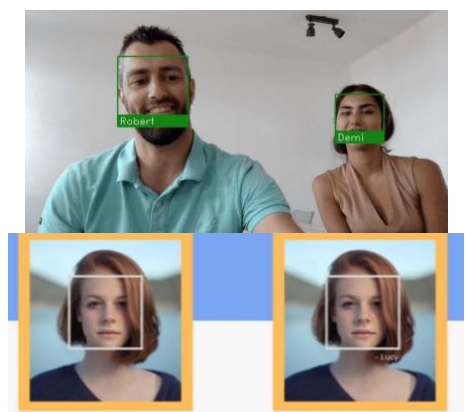

Fig. 3. Face detection in python

*C.* **Weapon detection and alarm trigging system -**

We detect the weapons hidden in the body or in the clothes depending on how their form shifts, are being worn. We are putting a lot of effort into finding hidden weapons and have already created a method to detect guns.

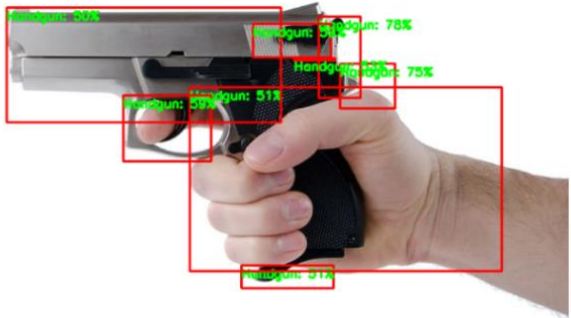

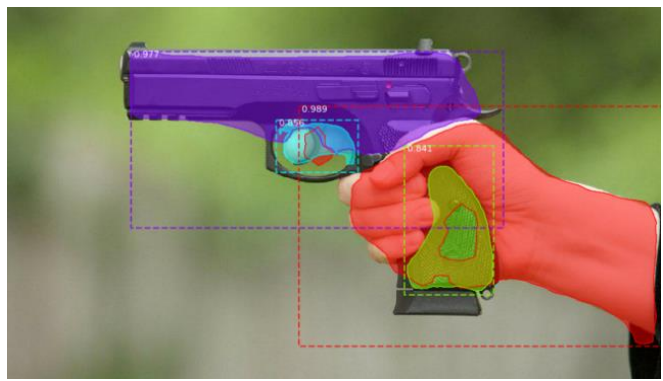

Fig. 4. Weapon detection in python

*D.* **Smart notification and content summarizer -**

Police officers may respond right away after receiving the notification thanks to our online application. When a weapon is detected, the system sounds the alert and notifies the user via the website. The notice contains all the pertinent details about the occurrence. the time, place, and date. For greater accuracy, it's conceivable that we'll submit a CCTV video and show it alongside the notification.

The police report or record is summarized on the other side. We all are aware that a police report or criminal report has lengthy paragraphs and circumstances. As a result, reading the record and acquiring a sense of the occurrence takes a lot of time for the officers under this approach, when a police officer enters any police report or document into the system, it summaries the extensive context and outputs the most significant and helpful lines or ideas to the officer. Officers can therefore quickly read the written summary and understand it. They only need to spend a short amount of time on this. In order to find the most challenging facts in documents, this study illustrates a programmed method for reflecting all accessible information using non-fuzzy ideas on a range of extracted data. This is highly unique to our study project since the recommended method, which employs a small number of fuzzy rules, may evolve, and be used with emerging intelligent robots that can assess writing.

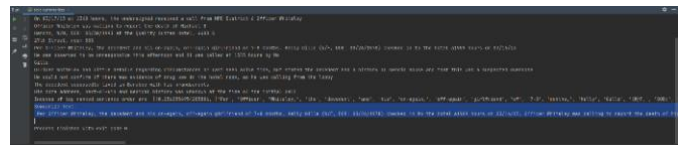

Fig. 5. Content summarizer

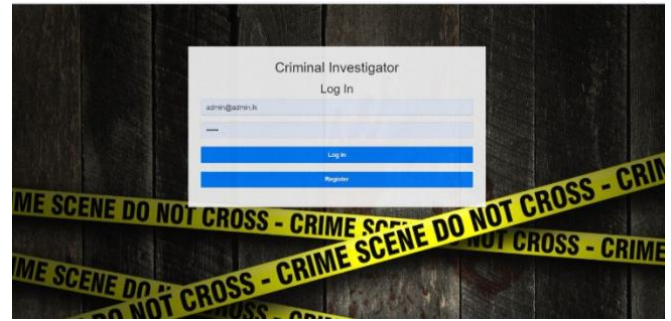

Fig. 6. Web application

## V. CONCLUSION AND FUTURE WORK

The recommended remedy allows for just incident detection. Though it would be possible to adapt this system to foresee incidents as technology advances. The data acquired from the CCTV stream will be used in the next system training. To provide a viable channel of communication between a crime or incident and pertinent agencies, the recommended web application should be enhanced further.

This study suggests a unique method for character training, crime including weapons, crime prediction based on real-world data, and face and fingerprint identification. The suggested approach was tested, and the results of the subsystems were compared to find confirmation. The police report summary is also given a lot of weight by the system.

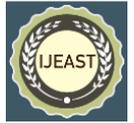